# Similarity Networks for the Construction of Multiple-Fault Belief Networks[*]

**David Heckerman**
Medical Computer Science Group
Room 215 MSOB, Stanford Medical Center
Stanford, California 94305

## Abstract

A similarity network is a tool for constructing belief networks for the diagnosis of a single fault. In this paper, we examine modifications to the similarity-network representation that facilitate the construction of belief networks for the diagnosis of multiple coexisting faults.

## 1 Introduction

A *similarity network* is a graphical extension of the belief-network representation [4, 5]. The representation is a tool for constructing large and complex belief networks of the form shown in Figure 1. In this belief network, the chance node FAULT contains many mutually exclusive and exhaustive instances. This node conditions many other nodes, but is itself not conditioned by any nodes. Belief networks of this form are seen commonly in problems of diagnosis in which a single fault is present.

The similarity-network representation was instrumental in the construction of the belief network for Pathfinder, a normative expert system for the diagnosis of lymph-node diseases [6, 7]. In Pathfinder's domain, the assumption that faults—the diseases—are mutually exclusive is appropriate, because co-occurring diseases almost always appear in different lymph nodes or in different regions of the same lymph node. In many domains, however, such an assumption is invalid. Patients admitted to the internal-medicine ward of a hospital, for example, often present with four to five coexisting diseases. In this paper, we examine how we can use the similarity-network representations to facilitate the construction of belief networks for the diagnosis of multiple faults.

[*]This work was supported by the National Library of Medicine under Grant RO1LM04529, and by the National Science Foundation under Grant IRI-8703710.

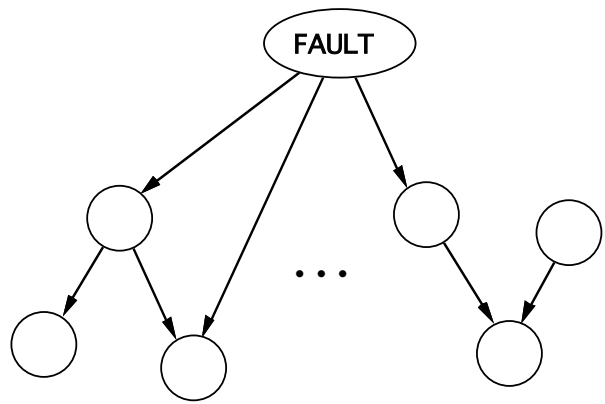

Figure 1: A belief network for the diagnosis of a single fault.

## 2 Similarity Networks

In this section, we use the similarity-network representation to construct a belief network for a small problem in medical diagnosis. The purpose of this exercise is to illustrate the basic concepts and techniques underlying this representation, and to demonstrate some of the advantages of its use. Because the example is small, however, the full power of this representation for simplifying knowledge acquisition cannot be demonstrated. In [5], I describe highlights of this approach to knowledge acquisition applied to Pathfinder. There, the power of these representations is illustrated more fully.

The medical example that we examine is real, but it has been simplified for purposes of presentation. Dr. Harold Lehmann served as the expert for the domain. The figures in the example were generated by SimNet, an implementation on the Macintosh computer of the similarity-network and partition representations. Readers interested in the details of SimNet's operation should consult [5].

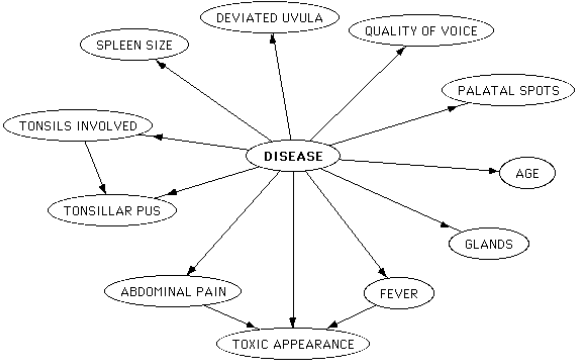

Figure 2: A belief network for sore throat.

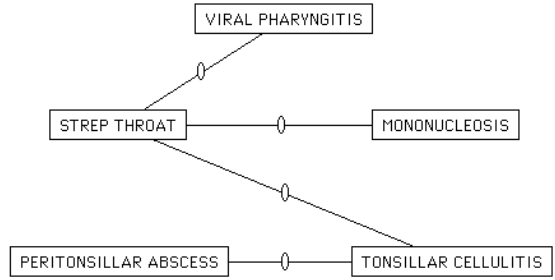

Figure 3: A similarity graph for sore throat. Although the graph is a tree (i.e., there is exactly one path between any two nodes in the graph), in general, similarity graphs can contain cycles.

Throughout the example, I will distinguish between the construction of a belief network, similarity network, or partition by a *person* and the construction of these representations by an *algorithm*. Consequently, the terms *to compose* and *to construct* will refer to situations where a person and an algorithm generate a representation, respectively.

Suppose a patient between 5 and 18 years of age comes to an emergency room complaining of severe sore throat. A belief network for this situation is illustrated in Figure 2. The chance node DISEASE represents the causes of sore throat: VIRAL PHARYNGITIS, STREP THROAT, MONONUCLEOSIS, TONSILLAR CELLULITIS, and PERITONSILLAR ABSCESS. Here, we assume that these diseases are mutually exclusive and exhaustive. The remaining nodes represent evidence relevant to the diagnosis of the patient's disease. We now discuss how to construct this belief network using a similarity network.

The focus for the composition of the similarity network is the *distinguished node* or *distinguished variable*. For medical domains, the distinguished variable represents a set of mutually exclusive and exhaustive diseases. In general, we refer to the mutually exclusive exhaustive instances of this variable as faults.

A similarity network consists of a *similarity graph* and a collection of *local belief networks*. To compose a similarity graph, we first compose the similarity graph. The nodes in the similarity graph correspond to individual faults. Informally, the edges in the similarity graph connect faults that are similar. We shall discuss soon the precise meaning of edges in a similarity graph. The similarity graph for sore throat is shown in Figure 3.

Next, we compose a *local belief network* for each pair of faults that is connected in the similarity graph. To compose a local belief network for the fault pair $f_i$ and $f_j$,[1] we imagine that one of these two faults has occurred. Given this supposition, we compose a belief network consisting of the distinguished node—whose instances are restricted to $f_i$ and $f_j$—and those *nondistinguished nodes* that are relevant to the discrimination of these faults. Formally, we omit a node from the local belief network if and only if the node would be disconnected from the distinguished node (i.e., there would be no path between the node and the distinguished node) if we included it in the network. The distinguished node must have no predecessors in a local belief network.

Figure 4(a) shows the local belief network for the edge between PERITONSILLAR ABSCESS and TONSILLAR CELLULITIS in the similarity graph. The small oval at the top of the belief network represents the distinguished node whose instances are restricted to these two diseases. The remaining nodes in the local belief network represent the features or disease manifestations that are relevant to the discrimination the diseases PERITONSILLAR ABSCESS and TONSILLAR CELLULITIS. Notice that there are no arcs among the nondistinguished nodes. The missing arcs represent the assertion that, given that the patient has either PERITONSILLAR ABSCESS or TONSILLAR CELLULITIS, all manifestations in the belief network are independent. Also note that there are fewer manifestations in this local belief network than in the belief network for the entire domain (Figure 2). This observation tends to be true, in general, because the diseases associated with local belief networks are similar.

---

[1] In this paper we use lowercase letters to denote variables and uppercase letters to denote sets of variables. We subscript a variable (e.g., $x_i$) to denote an instance of that variable. Similarly, we subscript a set of variables (e.g., $X_i$) to denote an instance of that set.

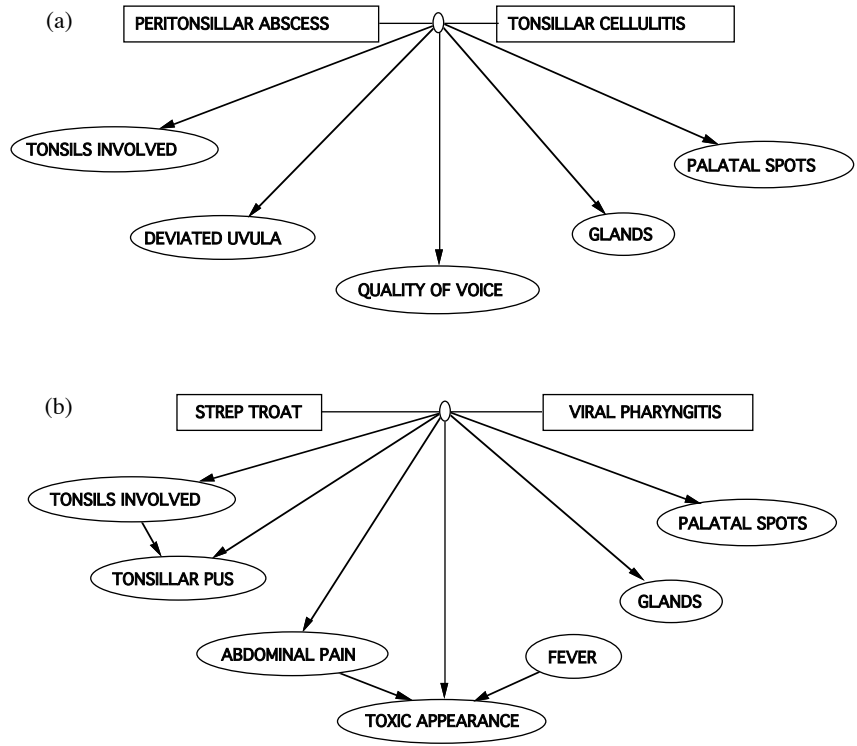

Figure 4: Two local belief networks. The small oval at the top of the belief network represents the distinguished node.

Figure 4(b) shows the local belief network for the edge between STREP THROAT and VIRAL PHARYNGITIS in the similarity graph. Again, the belief network contains fewer features than does the belief network for the sore-throat domain as a whole. Now, however, some of the disease manifestations are conditionally dependent. The arc from TONSILS INVOLVED to TONSILLAR PUS reflects the expert's assertion that the probability of seeing pus on a patient's tonsils depends on whether the disease involves one tonsil, both tonsils, or neither tonsil. The arcs from FEVER and ABDOMINAL PAIN to TOXIC APPEARANCE reflect the observation that a patient is more likely to present with a toxic appearance if the patient has abdominal pain or a high fever. Although FEVER is relevant to the discrimination of STREP THROAT and VIRAL PHARYNGITIS indirectly through its effect on TOXIC APPEARANCE, the missing arc from the disease node to FEVER represents the belief that temperature alone is not relevant to the discrimination of the two diseases.

Given the similarity network that we have *composed,* we can now *construct* the belief network for the full sore-throat problem, called the *global belief network*. Specifically, we construct the global belief network by forming the *graph union* of the local belief networks in the similarity network. The operation of graph union is straightforward. The nodes in the graph union of a set of graphs is the simple union of the nodes in the individual graphs. Similarly, the arcs in the graph union of a set of graph is the simple union of the arcs in the individual graphs. That is, a node (or arc) appears in the graph union, if and only if there is such a node (or arc) in at least one of the individual graphs.

The global belief network for sore throat was shown in Figure 2. The node QUALITY OF VOICE, for example, appears in the global belief network because it appears in the local belief network for PERITONSILLAR ABSCESS and TONSILLAR CELLULITIS. The arc from DISEASE to ABDOMINAL PAIN appears in the global belief network because it is present in the local belief network for STREP THROAT and VIRAL PHARYNGITIS.

If (1) the global belief network contains no directed cycles, (2) the joint probability distribution for the distinguished and nondistinguished variables is strictly positive (i.e., there are no probabilities in the distribution that are equal to zero), and (3) the similarity graph is connected (i.e., there is a path between any two nodes in the graph), then the construction of the global belief network from a simi-larity network is *sound* [5]. That is, any joint distribution that satisfies the assertions of conditional independence implied by the local belief networks also satisfies the assertions of conditional independence implied by the global belief network.

The formal criteria for drawing edges in a similarity graph are that we connect two diseases only if we can compose a local belief network for the disease pair, and the similarity graph must be connected. There is no formal requirement that connected diseases be similar. As we have seen in the medical example, however, local belief networks for pairs of similar diseases tend to exclude many of the features that are relevant to the set of diseases as a whole. Furthermore, by composing local belief networks for pairs of similar diseases, the expert can use a similarity network to focus his attention on precisely those diagnostic subproblems with which he is familiar. Thus, an expert can simplify greatly his task of composing the local belief networks by connecting only similar diseases in the similarity graph (provided the graph remains connected).

We can extend the similarity-network representation to include local belief networks for fault sets of arbitrary size. In such an extension, we replace the similarity graph with a similarity hypergraph. A *hypergraph* consists of nodes and *hyperedges* that connect sets of nodes. We then compose one local belief network for each hyperedge. To ensure that the global belief network constructed from such a similarity network is sound, we must replace only the constraint that the similarity graph be connected, using instead the constraint that the similarity hypergraph be connected.

A similarity network derives its power from its ability to represent assertions of conditional independence that are not conveniently represented in an ordinary belief network. To illustrate such an assertion, let $f_\subseteq$ denote a proper subset of faults. If $f$ and feature $x$ are independent, given that one of the elements of $f_\subseteq$ is present, we say that $x$ is not relevant to $f_\subseteq$. Formally, a feature $x$ *is not relevant to* the set $f_\subseteq$, given background knowledge $\xi$, if and only if

$$p(f_i | x_k, f_\subseteq, \xi) = p(f_i | f_\subseteq, \xi) \qquad (1)$$

for all instances $x_k$ of variable $x$, and for all faults $f_i$ in $f_\subseteq$. In Equation 1, the set $f_\subseteq$, which conditions both probabilities, denotes the disjunction of its elements. We call the form of conditional independence represented by Equation 1 *subset independence*. Using Bayes' theorem, we can derive an equivalent criterion for subset independence. In particular, we can show that a feature $x$ is not relevant to the set of faults $f_\subseteq$, given background knowledge $\xi$, if and

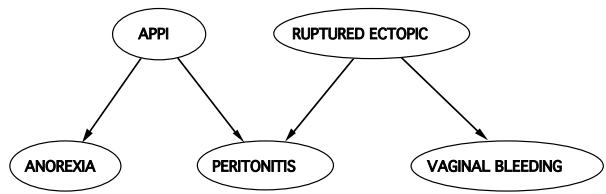

Figure 5: A belief network for the diagnosis of multiple diseases.

only if

$$p(x_k | f_i, \xi) = p(x_k | f_j, \xi) \qquad (2)$$

for all pairs $f_i, f_j \in f_\subseteq$, and for all instances $x_k$ of feature $x$.

Assertions of subset independence are *asymmetric*. In general, an assertion of conditional independence is asymmetric if it holds for only some instances of its variables. Assertions of subset independence, in particular, hold for only proper subsets of the disease variable $f$.

We cannot easily encode subset independence or other forms of asymmetric conditional independence in an ordinary belief network [5]. In contrast, such assertions are represented naturally by local belief networks. In particular, if we omit the feature $x$ from the local belief network for the faults $f_i$ and $f_j$, then we are asserting that $x$ is not relevant to the set $\{f_i, f_j\}$. In [5], I show how we can use these assertions of conditional independence to facilitate the assessment of the probability distributions associated with nodes in a belief network.

## 3 Belief Networks for Multiple-Fault Diagnosis

Let us now consider the problem of multiple-fault diagnosis. Figure 5 contains a small portion of a belief network for internal medicine. In this belief network, the node APPI represents the absence or presence of unreturned acute appendicitis. Similarly, the node RUPTURED ECTOPIC represents the absence or presence of acute ruptured ectopic pregnancy. Thus, this belief network does not exclude the possibility that both diseases can manifest in the same patient.

This example illustrates a difficulty that arises typically in situations where multiple faults are possible. In particular, both diseases in Figure 5 condition the node PERITONITIS, which represents the absence or presence of an inflammatory response in the peritoneum (the lining of the abdominal cavity). Thus, without any additional information, we would have to assess four probability distributions for this manifestation. More generally, we can have the situation, illustrated in Figure 6, where diseases $d^1, d^2, \ldots d^n$ can each cause manifestation $m$ to appear. Here, the node $m$ is associated with $2^n$ probability distributions.

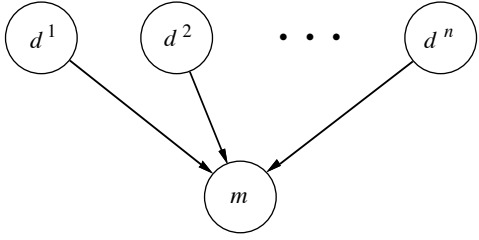

Figure 6: Multiple causes of the same manifestation.

## 4 Assumption of Causal Independence

We can reduce dramatically the number of probability assessments for node $m$ by making an additional assertion of conditional independence, called *causal independence*. In the context of Figure 6, let $p^i$ denote the probability that a patient, initially without disease $d^i$ and without manifestation $m$, will develop manifestation $m$ when getting disease $d^i$. When we assert causal independence in this situation, we state that probability $p^i$ does not depend on whether or not the patient has any other diseases before he has $d^i$, and that the manifestation $m$ cannot disappear when the disease $d^i$ manifests in the patient.

Now let $D_a$ denote an arbitrary instance of the set of variables $D = \{d^1, d^2, \ldots, d^n\}$. That is, let $D_a$ denote some assignment of absent or present to each disease $d^i$. In addition, let $D_-$ denote the particular instance of $D$ where all diseases are absent. Given the assertion of causal independence, the manifestation $m$ will be absent in a patient only if two conditions are met: (1) the manifestation $m$ cannot be present in the patient initially, and (2) none of the patient's diseases can act to cause $m$ to appear. Thus, we obtain

$$p(m_-|D_a, \xi) = p(m_-|D_-, \xi) \prod_{i \in \mathcal{I}_a} [1 - p^i] \quad (3)$$

where $m_-$ denotes the absence of manifestation $m$, and where $\mathcal{I}_a$ is the set of indices $i$ such that $d^i$ is present in $D_a$. Applying the sum rule to Equation 3, we obtain

$$p(m_+|D_a, \xi) = 1 - [1 - p(m_+|D_-, \xi)] \prod_{i \in \mathcal{I}_a} [1 - p^i] \quad (4)$$

where $m_+$ denotes the presence of manifestation $m$. If the patient has only disease $d^i$, Equation 4 becomes

$$p(m_+|\text{only } d^i_+, \xi) = 1 - [1 - p(m_+|D_-, \xi)] [1 - p^i] \quad (5)$$

where "only $d^i_+$" refers to the instance of $D$ where only $d^i$ is present. Solving for $p^i$ in Equation 5, and substituting the result in Equation 4, we obtain

$$p(m_+|D_a, \xi) = 1 - [1 - p(m_+|D_-, \xi)] \quad (6)$$
$$\cdot \prod_{i \in \mathcal{I}_a} \left[ \frac{1 - p(m_+|\text{only } d^i_+, \xi)}{1 - p(m_+|D_-, \xi)} \right]$$

Thus, with the assertion of causal independence, we can determine all the probability distributions associated with the node $m$ in Figure 6, from only the probabilities $p(m_+|D_-, \xi)$ and $p(m_+|\text{only } d^i_+, \xi)$, $i = 1, 2, \ldots, n$.

Good and other theorists have described various forms of the causal-independence assertion [1, 11, 10]. Pearl refers to the particular form we have discussed, where diseases and features are binary, as a *noisy OR-gate* [10]. (We consider the origin of this name in the following paragraph.) Several researchers have noted that we can apply the noisy OR-gate and more general forms of causal independence to numerous situations within domains ranging from medicine to motorcycle repair [2, 3, 9, 8].

The model of the noisy OR-gate, as we have examined it so far, makes reference to the appearance of diseases over time. We can also represent the model in a belief network, without such a reference, as illustrated in Figure 7. In the figure, the node labeled $d^i$–causes–$m$ represents the absence or presence of an intermediate event through which $d^i$ causes manifestation $m$ to be present with certainty. As is indicated by the label OR above the deterministic node $m$, if any of these intermediate events occur, then the manifestation $m$ will appear for certain (hence the name noisy OR-gate). The arc from $d^i$ to $d^i$–causes–$m$ reflects the assertion that the absence or presence of $d^i$ influences the probability distribution for the variable $d^i$–causes–$m$.[2] In particular, we assume that, if $d^i$ is absent, then the disease cannot act to cause $m$, whereas if $d^i$ is present, then it causes $m$ to be present with some probability greater than 0. This probability corresponds to $p^i$ in the temporal formulation of the model. The lack of arcs between nodes in the upper two rows of the influence reflects the assertion of causal independence. In particular, the missing arcs represent the statement that the probability distribution for the variable $d^i$–causes–$m$ depends neither on the absence or presence of any other disease nor on the absence or presence of any other event leading to the occurrence of $m$. We require the node $d^0$–causes–$m$ to capture the possibility that manifestation $m$ will appear when all diseases are absent.

---
[2]Here, we assume that the belief network is minimal.

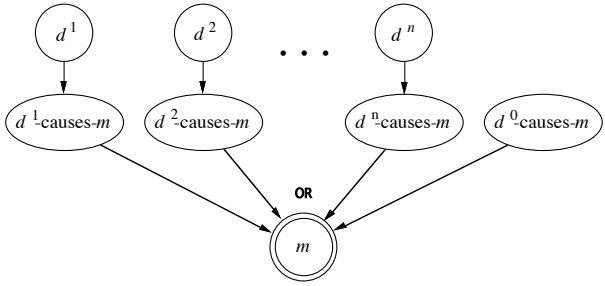

Figure 7: An atemporal representation of causal independence.

To make this model more concrete, let us consider the simple medical example in Figure 5. In this example, acute ruptured ectopic pregnancy can cause peritonitis, because blood from the rupture of a fallopian tube can collect in the peritoneal cavity, and thereby irritate the peritoneum. In contrast, the presence of unreturned acute appendicitis is associated with the release of substances that mediate the inflammatory response within the appendix. These substances can leak out of the appendix, and thereby cause an inflammatory response in the nearby peritoneum. Thus, the variable RUPTURED ECTOPIC–causes–PERITONITIS refers to the absence or presence of blood in the peritoneal cavity, whereas the variable APPI–causes–PERITONITIS refers to the absence or presence of inflammatory triggers of appendiceal origin in the peritoneum. To a good approximation, the probability that blood will collect in the peritoneal cavity is influenced neither by the presence of an unreturned acute appendicitis nor by the presence of inflammatory triggers of appendiceal origin in the peritoneum. Conversely, the probability that inflammatory triggers from the appendix will reach the peritoneum is influenced neither by the presence of an acute ruptured ectopic pregnancy nor by the presence of blood in the peritoneal cavity. Thus, we can assert causal independence for the interaction among these variables.

We can derive Equation 7 from both the temporal and atemporal models for causal independence. The atemporal model is somewhat problematic, because we often cannot define events of the form $d^i$–causes–$m$ precisely. Nonetheless, most people find this model easy to understand. In addition, we can use the framework to extend causal independence to situations where diseases and manifestations are not binary [3, 8].

## 5 Construction of a Multiple-Fault Belief Network

Now let us examine how we can use assumptions of causal independence in conjunction with an assessed similarity network to construct a belief network for the diagnosis of multiple diseases (or faults). The construction derives from Equation 7, which states that the only probability assessments we need to define the interaction illustrated in Figure 6 are those probabilities of the form $p\left(m_+|\text{only } d^i_+, \xi\right)$, $i = 1, 2, \ldots, n$ and the probability assessment $p\left(m_+|D_-, \xi\right)$. These probabilities are exactly those assessments that we can derive from a similarity network where we represent each disease as an instance of the distinguished node, and where we include NORMAL to represent the instance $D_-$.

With this observation in mind, let us consider the similarity network shown in Figure 8(a). From this similarity network, we can construct the multiple-disease belief network shown in Figure 5, in the following steps. First, we construct the global belief network from the similarity network, and transfer the manifestations ANOREXIA, PERITONITIS, and VAGINAL BLEEDING in the global belief network to the multiple-disease belief network. Also, if there were any arcs between these manifestations, we would transfer those arcs to the multiple-disease belief network. Second, for each node in the similarity graph, except NORMAL, we construct a binary chance node in the multiple-disease belief network. In particular, we construct the binary nodes APPI and RUPTURED ECTOPIC. Third, in the multiple-disease belief network, we draw an arc from APPI to ANOREXIA, and from APPI to PERITONITIS. Conversely, we do not draw an arc from APPI to VAGINAL BLEEDING. We can omit this arc because the local belief network for APPI and NORMAL states that the probability distribution for VAGINAL BLEEDING given APPI is equal to the distribution for VAGINAL BLEEDING given NORMAL, and because we assert causal independence. Similarly, we draw arcs from RUPTURED ECTOPIC to PERITONITIS and to VAGI-

NAL BLEEDING, but we do not draw an arc from RUPTURED ECTOPIC to ANOREXIA. Fourth, we use the probability assessments associated with the similarity network in conjunction with the noisy-OR-gate model (Equation 7) to compute the probability distributions for each manifestation. Finally, we assert that APPI and RUPTURED ECTOPIC are marginally independent, and assess the prior probabilities for these variables.

In transforming the similarity network to a multiple-disease belief network, we added several assertions of conditional independence. In particular, the similarity network implies only that the manifestations are conditionally independent given NORMAL, APPI alone, and RUPTURED ECTOPIC alone. The multiple-disease belief network, however, also encodes the assertion that the manifestations are independent given that both APPI and RUPTURED ECTOPIC are present in a patient. In general, when we apply the transformation described in the previous paragraph, we must verify that these additional assertions are valid.

Also, in transforming the similarity network to a multiple-disease belief network, we used the fact that NORMAL was connected to each of the remaining diseases in the similarity graph. That is, we used the fact that the similarity graph had a *star topology*, with NORMAL as its center. To understand this observation, let us consider the similarity network in Figure 8(b). Here, APPI and NORMAL are not connected, and thus we cannot identify those manifestations that are influenced by the chance node APPI in the multiple-disease belief network. Of course, we could add an arc from APPI to every manifestation in the multiple-disease belief network, but, in so doing, we would loose the assertions of conditional independence implied by the absence of the arc from APPI to ANOREXIA.

Although the transformation is facilitated by a similarity graph with a star topology, we should not require an expert to compose such graphs. Indeed, an expert might not be able to compose a local belief network for discriminating a particular disease from NORMAL. Fortunately, however, we can transform any similarity network to one whose similarity graph has a star topology. Specifically, given any similarity network, we first construct and assess the global belief network associated with that similarity network. Then, for each fault in the similarity graph (other than NORMAL), we construct a local belief network for discriminating that fault with NORMAL, using the probability distributions from the global belief network, and any ordering over the nondistinguished variables that is consistent with the global belief network (see [5, Chapter 3]). Once we obtain this new similarity network, we can construct the multiple-fault belief network from that similarity network as described previously.

## 6 General Algorithm

Let us now consider a general algorithm for transforming a similarity network $\mathcal{S}$ into a multiple-fault belief network $\mathcal{M}$. Let $f_0, f_1, f_2, \ldots, f_n$ denote the instances of the distinguished variable $f$ in $\mathcal{S}$, where $f_0$ is NORMAL. In addition, let $f^i, i = 1, 2, \ldots, n$, denote the binary variable in $\mathcal{M}$ that corresponds to fault $f_i$ in $\mathcal{S}$. Also, let $x$ and $y$ denote arbitrary nondistinguished variables in $\mathcal{S}$. Finally, let $\mathcal{S}'$ denote a similarity network constructed from $\mathcal{S}$ such that the similarity graph of $\mathcal{S}'$ has a star topology with center $f_0$, and let $\mathcal{G}_{\mathcal{S}'}$ denote the global belief network constructed from $\mathcal{S}'$. We transform $\mathcal{S}$ into $\mathcal{M}$ as follows ($x \longrightarrow y$ denotes an arc from $x$ to $y$):

> Construct $\mathcal{S}'$ from $\mathcal{S}$
> Construct $\mathcal{G}_{\mathcal{S}'}$ from $\mathcal{S}'$
> For all $x$ in $\mathcal{G}_{\mathcal{S}'}$, place $x$ in $\mathcal{M}$
> For all $x \longrightarrow y$ in $\mathcal{G}_{\mathcal{S}'}$, place $x \longrightarrow y$ in $\mathcal{M}$
> For $f_i$ in $\mathcal{S}'$, $i = 1, 2, \ldots, n$, place $f^i$ in $\mathcal{M}$
> For local belief network $f_{0i}$ in $\mathcal{S}'$, $i = 1, 2, \ldots, n$,
>     For all $x$ in $\mathcal{G}_{\mathcal{S}'}$, if $x$ is in $f_{0i}$, then place
>     $h^i \longrightarrow x$ in $\mathcal{M}$
> For all $x$,
>     Determine the probability distributions for $x$
>     in $\mathcal{M}$ from the distributions for $x$ in
>     $\mathcal{G}_{\mathcal{S}'}$, using assertions of causal independence
> Assess dependencies among the $f^i$ in $\mathcal{M}$
> For $f^i$ in $\mathcal{M}$, $i = 1, 2, \ldots, n$, assess the
> probability distributions for $f^i$

In this transformation, the nondistinguished variables do not need to be binary, provided we generalize beyond the noisy OR-gate model expressed by Equation 7. In addition, we can generalize this algorithm to include situations where $f^i$ is nonbinary, by representing each instance of $f^i$ in the similarity graph, and by identifying forms of causal independence that can account for the interaction between $f^i$ and each nondistinguished node.

Finally, note that we can use this algorithm to transform a *belief network* of the form shown in Figure 1 to a belief network for the diagnosis of multiple faults. As mentioned in the previous section, we construct $\mathcal{S}'$ from the global belief network derived from $\mathcal{S}$; we do not construct $\mathcal{S}'$ from $\mathcal{S}$ directly. Consequently, we do not require an original similarity network $\mathcal{S}$ for the transformation.

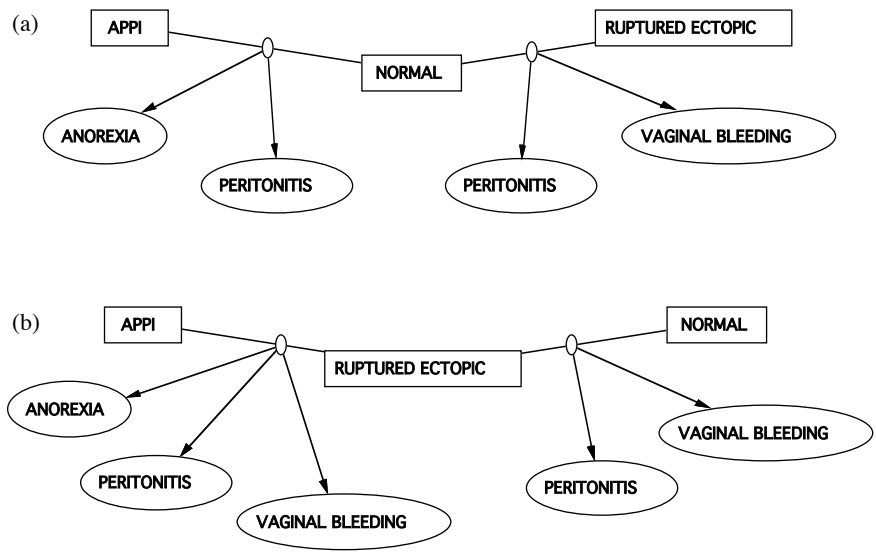

Figure 8: Two similarity networks for APPI and RUPTURED ECTOPIC.

## 7 Summary

The similarity-network representation was designed for the construction of belief networks for the diagnosis of a single fault. Nonetheless, in this paper, we have seen that we can also use this representation to facilitate the construction of belief networks for multiple faults.

## References

[1] I.J. Good. A causal calculus (I). *British Journal of Philoshophy of Science*, 11:305–318, 1961.

[2] J.D.F. Habbema. Models for diagnosis and detection of combinations of diseases. In F.T. de Dombal and F. Gremy, editors, *Decision Making and Medical Care*, pages 399–411. North-Holland, New York, 1976.

[3] D.E. Heckerman. Formalizing heuristic methods for reasoning with uncertainty. Technical Report KSL-88-07, Medical Computer Science Group, Section on Medical Informatics, Stanford University, Stanford, CA, May 1987.

[4] D.E. Heckerman. Probabilistic similarity networks. *Networks*, 20, 1990.

[5] D.E. Heckerman. *Probabilistic Similarity Networks*. PhD thesis, Program in Medical Information Sciences, Stanford University, Stanford, CA, June 1990.

[6] D.E. Heckerman, E.J. Horvitz, and B.N. Nathwani. Update on the Pathfinder project. In *Proceedings of the Thirteenth Symposium on Computer Applications in Medical Care,* Washington, DC, pages 203–207. IEEE Computer Society Press, Silver Spring, MD, November 1989.

[7] D.E. Heckerman, E.J. Horvitz, and B.N. Nathwani. Toward normative expert systems: The Pathfinder project. Technical Report KSL-90-08, Medical Computer Science Group, Section on Medical Informatics, Stanford University, Stanford, CA, February 1990.

[8] M. Henrion. Some practical issues in constructing belief networks. In L.N. Kanal, J.F. Lemmer, and T.S. Levitt, editors, *Uncertainty in Artificial Intelligence 3*, pages 161–174. North-Holland, New York, 1989.

[9] M. Henrion and D.R. Cooley. An experimental comparison of knowedge engineering for expert systems and for decision analysis. In *Proceedings AAAI-87 Sixth National Conference on Artificial Intelligence,* Seattle, WA, pages 471–476. Morgan Kaufmann, San Mateo, CA, July 1987.

[10] J. Pearl. *Probabilistic Reasoning in Intelligent Systems: Networks of Plausible Inference*. Morgan Kaufmann, San Mateo, CA, 1988.

[11] P. Suppes, editor. *A Probabilistic Theory of Causality*. North-Holland, New York, 1970.